\documentclass[conference,compsoc]{IEEEtran}
\usepackage{microtype}
\usepackage{blindtext}
\usepackage{array}
\usepackage[compact]{titlesec}
\titlespacing{\section}{1pt}{*1}{*1}
\titlespacing{\subsection}{1pt}{*1}{*0}
\titlespacing{\subsubsection}{1pt}{*0.5}{*0}
\usepackage{booktabs}
\usepackage{graphicx}
\usepackage{subfigure}
\usepackage{booktabs}
\usepackage{amsmath}
\usepackage{amsthm}
\usepackage{lipsum}
\usepackage{cleveref}
\usepackage{xcolor}
\usepackage{algorithm,algpseudocode}
\usepackage{graphicx}
\ifCLASSOPTIONcompsoc
  \usepackage[nocompress]{cite}
\else
  \usepackage{cite}
\fi
\newtheorem{lemma}{Lemma}

\begin{document}

\title{BAFFLE : Blockchain Based Aggregator Free Federated Learning}
\author{\IEEEauthorblockN{Paritosh Ramanan}
\IEEEauthorblockA{School of Industrial and Systems Engineering\\
Georgia Institute of Technology\\
Atlanta, Georgia\\
Email: paritoshpr@gatech.edu}
\and
\IEEEauthorblockN{Kiyoshi Nakayama}
\IEEEauthorblockA{TieSet Inc.\\
San Jose, California\\
Email: knakayama@tie-set.com}}
\maketitle

\begin{abstract}
A key aspect of Federated Learning (FL) is the requirement of a centralized aggregator to maintain and update the global model. However, in many cases orchestrating a centralized aggregator might be infeasible due to numerous operational constraints. In this paper, we introduce BAFFLE, an aggregator free, blockchain driven, FL environment that is inherently decentralized. BAFFLE leverages Smart Contracts (SC) to coordinate the round delineation, model aggregation and update tasks in FL. BAFFLE boosts computational performance by decomposing the global parameter space into distinct chunks followed by a score and bid strategy. In order to characterize the performance of BAFFLE, we conduct experiments on a private Ethereum network and use the centralized and aggregator driven methods as our benchmark. We show that BAFFLE significantly reduces the gas costs for FL on the blockchain as compared to a direct adaptation of the aggregator based method. Our results also show that BAFFLE achieves high scalability and computational efficiency while delivering similar accuracy as the benchmark methods.

\end{abstract}
\section{Introduction}
\label{introduction}
Federated Learning (FL) \cite{fl2} is a distributed machine learning paradigm that accomplishes large scale learning tasks \cite{sysmlfl19}. FL leverages data sets localized on end user devices in order to ensure privacy. A fundamental assumption of the FL paradigm is the presence of a centralized aggregator meant to coordinate the global computational progress. An aggregator discharges four main functions in the FL paradigm. First, it is responsible for delineating the global computational process into distinct rounds. Second, it maintains a global estimate of the machine learning model to be updated after every round. Third, the aggregator selects the end user devices and sends a copy of the global model estimate to each. Lastly, the aggregator performs the critical step of updating the global model estimate based on the local copies of selected user devices. 

The requirement of central aggregator raises operational challenges, especially in FL applications wherein the global model is of vital operational and diagnostic value to the end users \cite{roy2019braintorrent,passerat2019blockchain}. First, in many instances, implementing a central aggregator might not be a feasible option due to logistical challenges \cite{he2019central,roy2019braintorrent}. Second, end users must trust the aggregator's selection and update mechanism for the user devices and their local models respectively. In case an aggregator holds a bias towards specific users, the final global model might not generalize well \cite{mohri2019agnostic}. Third, the central aggregator results in a single point of failure for the FL task, thereby raising robustness concerns \cite{hegedHus2019gossip}. Lastly, the central aggregator is typically cloud based \cite{sysmlfl19,fl2}. Access to the cloud, might be out of reach for consortia of small industries due to skill and expertise requirements \cite{carcary2014adoption}. As a result, a central aggregator might induce a high barrier of entry for small organizations which might be incapable of implementing large scale FL tasks. 

Blockchain based decentralization can be effectively leveraged for alleviating operational issues with respect to a centralized aggregator.  However, careful consideration of the computational constraints imposed by the blockchain is required in order to realize an aggregator free FL scheme\cite{harris2019decentralized}. First, storage of data and computation on the blockchain incurs significant costs. 
Second, pushing an entire machine learning model to the blockchain becomes computationally bulky potentially incurring heavy latency due to consensus. Lastly, there are limits on transaction size imposed by the blockchain protocols that restrict the amount of data that can be stored and updated on blockchain in a single transaction. These computational constraints place limitations on the model aggregation and update process in FL. Nevertheless, the need for a scalable, low cost framework that retains the benefits of aggregator free FL while fulfilling the computational constraints of the blockchain has so far not been addressed \cite{harris2019decentralized}.

In this paper, we propose BAFFLE, a blockchain based aggregator free FL environment. BAFFLE leverages SmartContracts (SCs) to maintain the global model copy and the associated computational state of the users. By its very design, BAFFLE enables users to update the global model on the SC independently and in parallel, leading to significantly lower computational costs. On the operational front, for a particular round, selection of end users in BAFFLE is based on the worth of their local updates as assessed by the SC. BAFFLE ensures that rounds are delineated according to the reported computational state of all the users thereby avoiding bias. Lastly, owing to a fully decentralized, agggregator free approach, BAFFLE saves on cloud setup and operational costs and eliminates technical expertise requirements for maintaining centralized aggregators\cite{cld2}. Therefore, BAFFLE is able to deliver high computational efficiency while successfully eliminating the operational limitations of an aggregator driven approach.

From a social standpoint, the computational benefits of BAFFLE coupled with elimination of cloud based costs and expertise requirements lowers the entry barrier for small organizations. BAFFLE can be used by micro scale organizations on public or private blockchains to self organize and leverage FL among their peers in a computationally friendly way. In doing so, each organization in a community can preserve their own data privacy but yet collaborate on building a global model that helps address challenges common to the entire community. As a result, BAFFLE can effectively be used to empower communities of users who would otherwise not have the capability to obtain robust machine learning models for their own internal challenges.

In this paper, we show that BAFFLE  is able to successfully circumvent operational constraints of FL and deliver its benefits in a computationally sound manner. 
BAFFLE consists of a budgeted approach towards the model update and aggregation steps and leverages SCs to delineate the rounds. We theoretically show that a classical FL scheme is equivalent to a BAFFLE driven approach with a linear relation between the respective learning rates. 
We provide a practical, production level implementation of BAFFLE on a private Ethereum network, with Solidity powered SC deployments. We demonstrate the merits of BAFFLE on a real world case study using a large Deep Neural Network(DNN) model. Based on our case study, we perform exhaustive experiments to study the user benefits, robustness and scalability of BAFFLE compared to other benchmarks. Our results indicate that BAFFLE provides superior computational performance despite the highly restrictive constraints imposed by the blockchain.




Our paper is organized as follows. In Section \ref{sec:rw} we provide an overview of related work pertaining to the fields of blockchain and decentralized ML. Section \ref{sec:sfl} discusses the novel strategies employed in BAFFLE to circumvent the restrictions imposed by the blockchain. Section \ref{sec:cpb} provides an overview of the local and global computational perspectives of BAFFLE.
Section \ref{sec:cst} introduces a real world case study of improving driver revenue where an aggregator free FL mechanism could be highly beneficial. Section \ref{sec:exp} deals with the entire set of experiments and their analysis. We conclude the paper in Section \ref{sec:conc} in addition to providing a quick overview of future work.

\section{Related Work}\label{sec:rw}
Improving a global neural network model using distributed data with a privacy-preserving purpose was first studied in \cite{shokri2015privacy}. The authors provide a scheme of jointly learning an accurate model by multiple parties for a given objective. More specifically, they consider a global shared memory model where parameters of the global model are held. Various agents participating in this framework can update a random subset of global parameters based on their local training. 

Federated Learning was later proposed in \cite{FL_Original, fl2} with its theoretical basis explored in \cite{konecny2015federated}. The authors provide an effective method for building collective knowledge across a set of devices while preserving their individual autonomy and privacy. More recently, there have been renewed efforts to scale up the FL framework as presented in \cite{sysmlfl19}. Such frameworks consider multiple aggregators headed by a master in order to manage the entire FL process. Although the work proposes a distributed network of aggregators coordinated by a master, it is not inherently aggregator free. 


Recent works \cite{lalithafully, lalitha2019peer} propose a framework of fully decentralized FL in which users update their belief by aggregating information from neighbors. While the theoretical aspect of decentralized FL is explored in these works, numerous system and architectural issues persist in achieving true decentralization. As a result, such systemic issues need to be dealt with in order to obtain a FL framework that is feasible under practical settings.

Practical efforts to integrate AI onto the blockchain are largely confined to white paper proposals without any tangible real world implementations available. The framework proposed in \cite{CortexWhitePaper} designs an SC based machine learning platform allowing users to upload tasks as well as contribute models to solve existing tasks. A distributed, AI computing platform has also been proposed in \cite{doornik1994practical} where mining nodes earn their income from processing AI models. 



There are also several projects that integrate federated learning into blockchain technologies. The work done in \cite{kim2018device} supports implementing the FL framework into the mining mechanism of the underlying blockchain platform. However, owing to modification requirements to the underlying consensus protocols such approaches tend to be cumbersome to implement on off the shelf blockchain platforms. The work done in \cite{harris2019decentralized} proposes and implements a decentralized AI framework using the blockchain. However, a key requirement of this framework is that training data from devices needs to be published on the blockchain. As a result, the data privacy benefits of FL paradigm is eliminated. In fact, the authors note that a decentralized, blockchain based AI framework with data privacy is a key component of their future work. 

Despite the above mentioned attempts, a concrete, practical framework for realizing decentralized aggregator free FL is so far lacking both in research and in industrial domains. To the best of our knowledge, BAFFLE is one of the first attempts at a production-level decentralized FL platform compatible with existing blockchains such as Ethereum. 

\section{Smart FL Contract Design: Decentralizing Role of Aggregator}\label{sec:sfl}
As mentioned in Section \ref{introduction}, a number of technical aspects need to be considered in order to make the FL process aggregator free. In this section, we examine the salient features of BAFFLE that allows us to circumvent blockchain based system constraints without compromising on solution quality. Even though BAFFLE has been implemented and evaluated on the Ethereum platform, the same technical principles would extend over to other blockchain based SC platforms as well.
\subsection{Chunking}
Most blockchain platforms have an upper limit pertaining to the data size of each transaction. 
For the Ethereum Virtual Machine (EVM) with the version we have used, this limit has been set to 24 kB by default. 
Such a limitation immediately results in a bottleneck for an aggregator free FL scheme since the underlying machine learning models are usually significantly larger than the block sizes. Such a system induced constraint necessitates the need for partitioning the machine learning model weight vector into numerous \textit{chunks} such that each chunk size is less than the maximum transaction size. However, chunking in turn introduces a few other notable aspects with regards to model sharing that can be described as follows.

\subsubsection{Serialization} 
Since storage on the SC is expensive, the machine learning model needs to be stored in a serialized format. However, partitioning the model after serialization could lead to inconsistencies. Therefore, for a specific FL task, it is important to first generate a partitioning scheme that must be used by all agents followed by individual serialization of the chunks. Such a \textit{chunk-and-serialize} scheme has numerous benefits. 
First, the chunks can be read to and written from independently and seamlessly. Second, such an independence among chunks can be exploited for parallel updates from multiple devices at the same time. Lastly, a chunk independence scheme also leads to a potential scoring technique wherein parts of the model can be evaluated for their worth. 

\subsubsection{Budgets} 
A potential benefit of chunking is that user devices are empowered to decide their levels of contribution individually. Since, pushing chunks on the blockchain involves a computational cost as well as miner fees, users can independently evaluate their own cost to benefit ratio and decide the number of chunks that they wish to update in a round. The maximum limit on the number of chunks that a user device wishes to update is referred to as the budget for that device. As a result the set of budget values from all user devices can be heterogenous in nature.

\subsection{Bidding and Round Delineation}
Each chunk is assigned a score by the end user devices themselves based on a norm difference with respect to the latest available global copy. Depending on a random selection the user device submits bids on a set of chunks as allowed by the budget limit. The SC receives bids on a diverse set of chunks from different user devices in every round. For chunks on which multiple bids were submitted, the device submitting the maximum score is chosen as the sole updater of the chunk.

Owing to the decentralized nature of our approach, the onus of delineating the rounds rests with the end user devices themselves aided by the information maintained on the SC. Specifically, a Participation Level (PL) is chosen for every FL learning task which specifies the number of agents which must have submitted their bids in order for the round to start. Once the participation level criteria is met, the round begins and no new devices are allowed to participate. Devices upload the chunks on whom their bids were accepted and proceed to signal a close of their round. 

\section{Computational Perspectives of BAFFLE}\label{sec:cpb}
The entire aggregator free blockchain based FL paradigm presented by BAFFLE can be viewed in terms of two important perspectives. The local perspective comprises of the computational steps undertaken by the user devices and their interaction with the blockchain. The global perspective details the SC driven scheme geared towards executing the various FL steps in an aggregator free setting.

\subsection{User Oriented Local Computation}
Locally, users interact with the blockchain through a client process that interfaces with the acquired user data. The client process is responsible for local training followed by leveraging the deployed SC in order to push the local model to the blockchain. 
Therefore, training and blockchain interaction form the two important aspects for every user device participating in BAFFLE. 
\subsubsection{Local Training}
User devices continuously observe new data points from their environment which can be leveraged for the FL task at hand. The user device pulls the latest available model from the blockchain through the SC. The globally acquired model is averaged with the latest available local copy. The resulting model is used to train on the locally available data to yield a new local model estimate. It is this new model estimate that now becomes a candidate for being pushed to the blockchain in the subsequent round.
\begin{figure}[!ht] 
 \centering
  \includegraphics[width=0.45\textwidth,keepaspectratio]{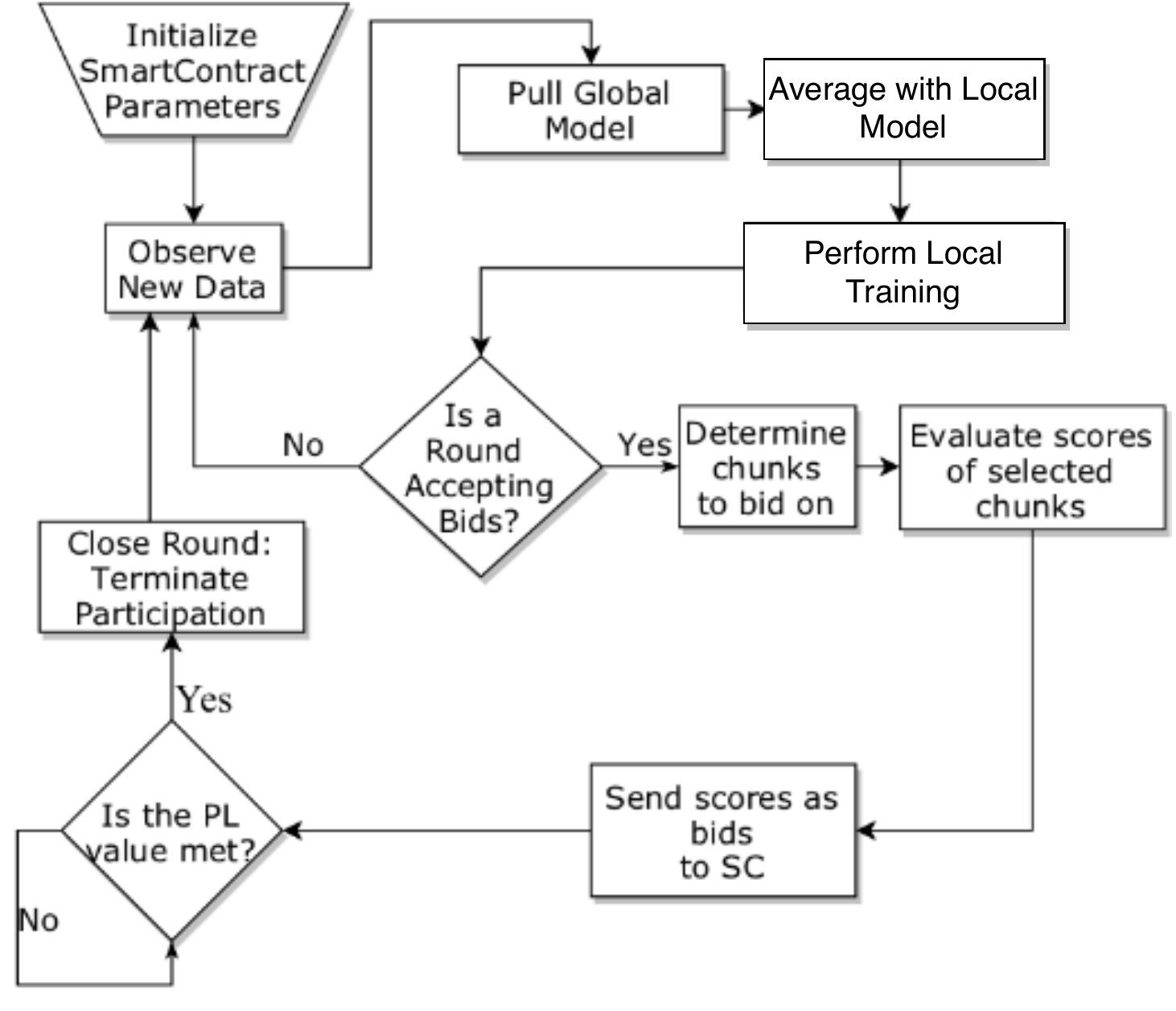}
  \caption{Flowchart depicting the sequence of events at the agent level}
  \label{fig:flowchart}
\end{figure}
\begin{algorithm}
\caption{Agent driven SC based update mechanism (on Agent $j$)}\label{alg:scalg}
\begin{algorithmic}
\State initialize partition scheme $\mathcal{C}$, local model $Q^j_0$
\For{$k=0 \ldots$}
\State obtain $Q^c, \forall c \in \mathcal{C}$ from blockchain using SC
\State compute $Q^{j,c}_{k+1} \leftarrow \frac{Q^c+Q^{j,c}_{k}}{2}, \forall c \in \mathcal{C}$ 
\State perform local training and update $Q^j_{k+1}$
\If{round is open for participation}
\State choose chunks $\tilde{C}^k \subseteq \mathcal{C}, |\tilde{C}^k| = B$ randomly
\State calculate scores $\delta_c = ||Q^c-Q^{j,c}_{k+1}||$, $\forall c \in \tilde{C}^k$
\State submit bids $[c,\delta_c], \forall c \in C^k$ to SC
\State determine accepted chunk set $C^k \subseteq \tilde{C}^k$ 
\State push $C^k$ to blockchain
\EndIf
\EndFor
\end{algorithmic}
\end{algorithm}
\subsubsection{Model Aggregation and Update}
In order to aggregate with the other devices and push its update to the chain, every agent considers the local model copy obtained after local training. The steps taken by the user device for model aggregation and update can be traced with the help of the flowchart depicted in Figure \ref{fig:flowchart} and summarized concisely in Algorithm \ref{alg:scalg}. Each user device is initialized on the basis of a fixed partition scheme denoted by set of chunks $\mathcal{C}$ with a maximum budget $B$. Let $Q^c,Q^{j,c}$ represent the estimate of parameters contained in chunk $c$ on the SC and agent $j$ respectively. For round $k$, user devices pull the global model copy, average with local copy, perform local training and check for the round status. In case a round is already underway and thus inactive, the user device returns to the task of collecting new data. If a round is active and accepting bids, devices choose randomly from their local chunks based on their budget size. A scalar score $\delta_c$ is assigned to each chunk based on the norm of the difference of the local weights with the global weights copy. These scores form the basis of the bid submitted to the SC which decides on which user device gets to update which chunk during round $k$. 
The probability of picking $B$ chunks including chunk $c$ at any device is given by ${C-1 \choose B-1 }/{C \choose B} = \frac{B}{C}$. In case of multiple devices bidding on the same chunk, we assume that the probability of winning is uniform. Therefore, using a binomial distribution we can say that in a particular round, the probability that device $j$ updates chunk $c$ is denoted by $\mu\cdot\frac{B}{C}$, where $\mu = \sum\limits_{d=0}^{L-1}\frac{1}{d+1}{L-1 \choose d}\Big(\frac{B}{C}\Big)^d\cdot\Big(1-\frac{B}{C}\Big)^{L-d-1}$ and L is the number of participants in the round (PL value).
\subsection{Globally Relevant Blockchain Driven FL}
\begin{figure*}[!ht]
\centering
  \includegraphics[width=0.9\textwidth,keepaspectratio]{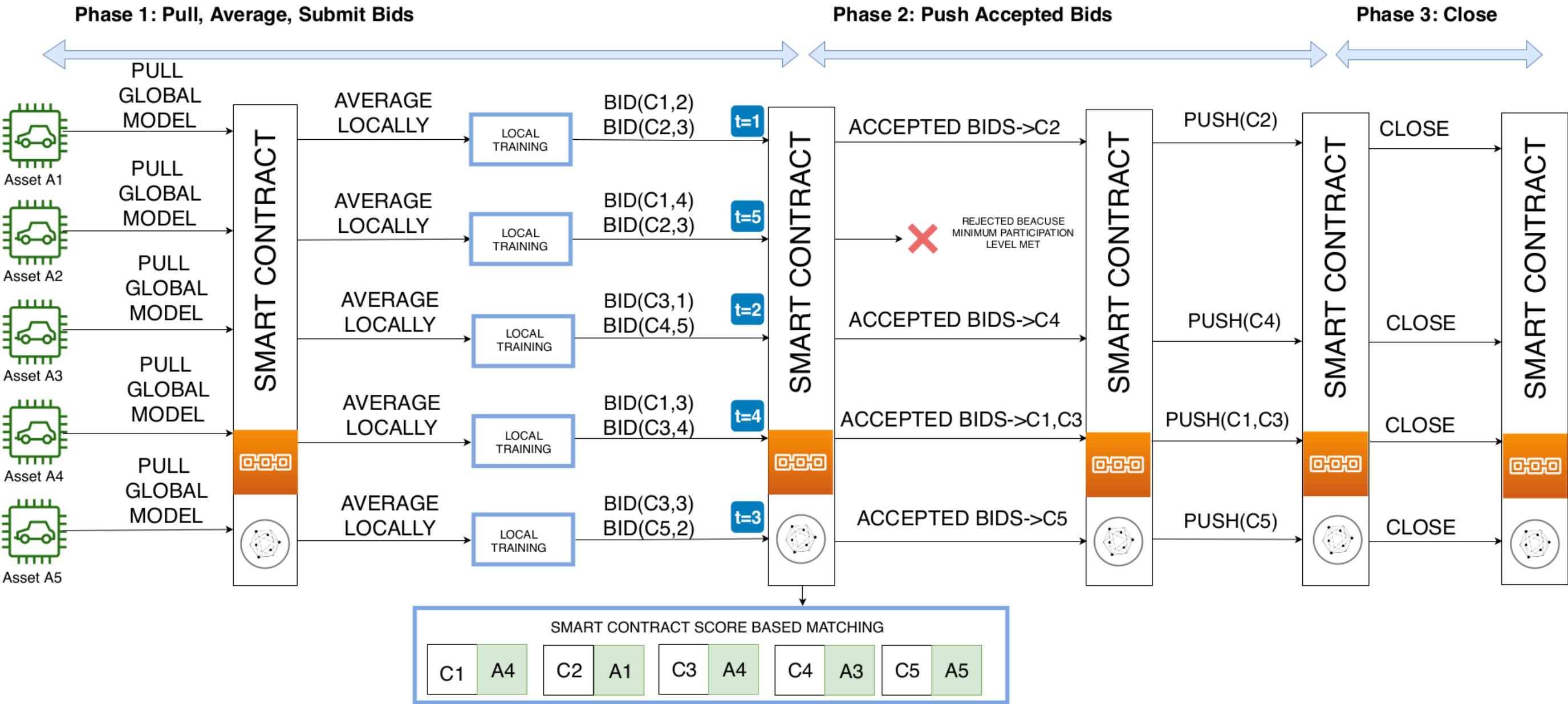}
  \caption{Global computational steps}
  \label{fig:overall_pic}
\end{figure*}
Globally, the computational process employed by BAFFLE is divided into three distinct phases. We illustrate the global computational perspective with the help of an example shown in in Figure \ref{fig:overall_pic}. In our example we consider a BAFFLE system comprising of 5 asset devices A1, A2, A3, A4, A5 respectively. The model is divided into 5 chunks C1, C2, C3, C4 and C5. For this example, we consider a PL value of 4. In Phase 1, each device performs local training and aggregation to generate new bids. Next, every device attempts to submit bids for its randomly chosen chunks. The bids chronologically arrive in the order A1, A3, A5, A4 and A2. In Phase 2, owing to the PL value being met with the arrival of bids from A4, A2 is rejected from the current round. The accepted devices push the respective chunks for which their bids were accepted. In Phase 3, every device eventually signals the culmination of all its local steps to the SC to mark the end of Phase 3 as well as the current round.

The SC responsible for the global aggregation scheme follows a contract-oriented design principle. Fields of significance contained in the SC are listed in Table \ref{tab:tab-scfeat}. For each attribute, we have set up an appropriate modifier that restricts the access to modify the value. In particular, the modifier for the AI Model data is designed only for the potential FL contributors that can benefit from the global model updated by the participants.
\begin{table*}[!htb]
\centering
\caption{SC Attributes in BAFFLE}
\label{tab:tab-scfeat}
\begin{tabular}{c|c}
\toprule
\textit{Attribute}   & \textit{Description}      \\ \hline
Model ID  &  Unique identifier assigned for every FL task by the SC\\ \hline
Round Registration Details & List of users with submitted bids for the upcoming round\\ \hline
Participation Level &  The minimum number of users with submitted bids required to begin a round\\ \hline
Chunk Core &  A data structure for every chunk holding: last updated time;\\
Array & last user to update; set of submitted scores \& their owners \\\hline
\bottomrule
\end{tabular}
\end{table*}

From a theoretical perspective, using Lemma \ref{lma1}, we show that the global computational process is equivalent in expectation to classical FL scheme with a learning rate scaled by a constant. 
\begin{lemma}\label{lma1}
Given $n$ devices, $C$ total chunks, $L\leq n$ agents per round, with budget $B\leq C$ the learning rate of BAFFLE ($\eta_{BFL}$) in expectation, can be denoted as:
\[\eta_{BFL}= \frac{2\cdot C\cdot\alpha_{FL}}{B\cdot\mu\cdot L\cdot\alpha_{BFL}}\cdot\eta_{FL}\]
where $\alpha_{FL},\alpha_{BFL}$ are the probabilities of agent selection at each round for classical FL and BAFFLE respectively. and $\eta_{FL}$ denote the learning rate of classical FL respectively.
\end{lemma}
\vspace{-2mm}
\begin{proof}
We know that the following relation holds for federated averaging \cite{mcmahan2017communication}:
\begin{equation}
    \hat{w}^{k+1}_{i} = \hat{w}^0_{i} - \eta\sum\limits_{t=1}^{k}\nabla f(\hat{w}^t)_i
\end{equation}
where $\hat{w}^k_i$ is the estimate of the $i^{th}$ component of the weight vector at round $k$, $\eta$ is the learning rate. Further, $\nabla f(\hat{w}^k)_i$ is $i^{th}$ component of the gradient estimated based on the globally available weight vector. 
In case of BAFFLE, we can say that 
\begin{subequations}
\begin{align}
    \hat{w}^{k+1}_{i} & =\frac{1}{2}[ \hat{w}^{k}_{i}+ \hat{w}^k_{i} - \eta_{BFL}\nabla f(\hat{w}^k)_i]\\
\hat{w}^{k+1}_{i} &  =\hat{w}^{k}_{i}- \frac{\eta_{BFL}}{2}\nabla f(\hat{w}^k)_i
\end{align}
\end{subequations}
Therefore, if at the $t^{th}$ round, device $j_t$ is active and the $i^{th}$ component is chosen, it follows that the expected value of the weight vector is given by:
\begin{equation}\label{eq:eq3}
    E[\hat{w}^{k+1}_i] = \hat{w}^0_i -\frac{\eta_{BFL}}{2} E\left[\sum\limits_{t=1}^{k}\nabla f_{j_t}(\hat{w}^t)_i\right]
\end{equation}
At every round, the probability of user device $j_t$ being selected is denoted by $\alpha_{BFL}$. Therefore, Equation \eqref{eq:eq3} is equivalent to:

\begin{equation}
E[\hat{w}^{k+1}_i]=\hat{w}^0_i -\frac{\eta_{BFL}}{2} \Bigg[\sum\limits_{t=1}^{k}\alpha_{BFL}\sum\limits_{j=1}^{n}\frac{B\cdot\mu\cdot\nabla f_{j_t}(\hat{w}^t)_i}{C}\Bigg]
\end{equation}
which leads to:
\begin{equation}\label{eq:eqBC}
    E[\hat{w}^{k+1}_i] = \hat{w}^0_i -\frac{B\cdot\mu\cdot\eta_{BFL}\cdot\alpha_{BFL}}{2\cdot C} E\Bigg[\sum\limits_{t=1}^{k}\sum\limits_{j=1}^{n}\nabla f_{j_t}(\hat{w}^t)_i\Bigg] 
\end{equation}
On the other hand, with aggregator driven FL, with $L$ user devices aggregated in each round, we can similarly state:
\begin{equation}\label{eq:eqFL}
E[\hat{w}^{k+1}_i] = \hat{w}^0_i - \frac{\eta_{FL}.\alpha_{FL}}{L}\Bigg[\sum\limits_{t=1}^{k}\sum\limits_{j=1}^{n}\nabla f_{j_t}(\hat{w}^t)_i\Bigg]
\end{equation}
Therefore, based on Equations \eqref{eq:eqBC} and \eqref{eq:eqFL}, we can say that the following relationship holds between $\eta_{BFL},\eta_{FL}$ in expectation.
\begin{equation}\label{eq:lmfinal}
\eta_{BFL}= \frac{2\cdot C\cdot\alpha_{FL}}{B\cdot\mu\cdot L\cdot\alpha_{BFL}}\eta_{FL}
\end{equation}
\end{proof}
\vspace{-3mm}
We now proceed to illustrate the benefits of BAFFLE on a real world case study involving the improvement of driver revenue for the taxi and ride sharing industry.
\section{Case Study: Improving Taxi Driver Revenue with BAFFLE}\label{sec:cst}
A key problem in the taxi and ride sharing industry is to improve driver revenue by reducing idle time \cite{han2016routing}. Drivers are often unable to find passengers at certain locations in the city at varying points of time during the day due to low demand \cite{han2016routing}. As a result, they usually hover around the same location until they find a passenger. Idling time reduces vehicle utilization and leads to potential loss in revenue for the individual driver \cite{markovTaxi}. 

The application of machine learning to improve driver revenue by reducing idle time has been studied before \cite{han2016routing,markovTaxi,taxiDRL}. Based on existing work, a Deep Reinforcement Learning (DRL) scheme is demonstrated to provide good quality improvement in driver revenues \cite{taxiDRL}. However, these approaches assume the presence of a centralized coordinator to steer the RL process. A central repository of ride information presents several privacy issues which have been successfully exploited to de-anonymize passenger information \cite{taxiPrivacy}. The work done in \cite{taxiDistDRL} as an extension of \cite{taxiDRL} introduces privacy preserving features and distributed computation as a means to improve driver revenue. However, \cite{taxiDistDRL} assumes a hierarchical computational setup that prevents all the benefits of decentralized computations from being realized in their entirety. The requirements of multiple control centres to perform the learning tasks leads to limited applicability of such approaches.

\subsection{Benefits of Aggregator Free FL for Improving Driver Revenue}
The taxi and ride sharing industry is a perfect example of micro scale enterprises that could benefit significantly from an aggregator free FL approach. The ride sharing and taxi industry remains largely an unorganized market where setting up a trusted coordinator remains a challenging proposition. Even in case of a central data repository, extracting intelligence from the anonymized data proves to be a futile exercise \cite{taxiPrivacy}. 
Moreover, drivers usually also do not have access to sophisticated computing platforms on which they could orchestrate learning tasks to improve their revenue. Therefore, a decentralized aggregator free FL environment allows drivers to leverage their collective ride experiences and improve their revenue without sharing their private ride data itself. 

\subsection{Deep Batch Reinforcement Learning for Taxis}
We use a batch DRL paradigm  to learn the Q function values and employ the Deep Neural Fitted Q \cite{nfq} method to accomplish our learning task. Specifically, we define our states and actions as follows:
\begin{itemize}
    \item Pickup State $s_i$\textit{:\textless pickup\_location, pickup\_time \textgreater}
    \item Dropoff State $s'_i$\textit{:\textless dropoff\_location, dropoff\_time \textgreater}
    \item Action \textit{a: action (dropoff\_location)}
    \item Reward \textit{r: fare}
\end{itemize}
State is defined by $S\times T$, where $S$ is set of discrete cells that divide the city into distinct grids. $T$ is set of 96 discrete intervals of 15 mins each for 24 hours. Therefore, given $N$ rides, we denote the ride set $\mathcal{H} = \{(s_i,a_i,s'_i,r_i),\forall i\in \{1,\ldots,N\}\}$. 
\begin{subequations}\label{eq:ql}
\begin{align}
\tilde{Q}_k(s_i,a_i) &= r_i + \gamma \max_{b}Q_k(s'_i,b), \forall i\in \mathcal{H} \label{eq:qbellman}\\
Q_{k+1} &\leftarrow \tilde{Q}_{k} - \eta \nabla\tilde{Q}(s_i,a_i)\label{eq:qsgd}
\end{align}
\end{subequations}
Equations \eqref{eq:qbellman} and \eqref{eq:qsgd} govern the functioning of the batch DRL framework at the $k^{th}$ round. The Q function is updated based on Equation \eqref{eq:qbellman} before being trained on the DNN using Equation \eqref{eq:qsgd}. 
\begin{algorithm}
\caption{BAFFLE for Improving Driver Revenues}\label{alg:mw}
\begin{algorithmic}
\For{taxi: $j=1\ldots P$}
\State initialize model $Q^j_0= Q^{init}$, budget $B$
\State initialize chunk set $\mathcal{C}$ based on given partition scheme.
\For{$k=0 \ldots$}
\State observe new ride set $\mathcal{H}^k$
\State pull latest available model $Q$ from blockchain
\State perform averaging $Q_k \leftarrow \frac{Q^j_k+Q}{2}$
\State update $\tilde{Q}^j_k$ based on Equation \eqref{eq:qbellman}
\State locally train $Q^j_{k+1}$ via Equation \eqref{eq:qsgd}
\State employ Algorithm \ref{alg:scalg} to push updates to SC
\EndFor
\EndFor
\end{algorithmic}
\end{algorithm}
In Algorithm \ref{alg:mw}, we consider $P$ taxis and begin by initializing all user devices to the same initial state. Next the partition information and SC details is loaded on each device. The user devices utilize a new set of rides accumulated locally in every round. The local estimate of the Q function is updated and trained locally based on Equations \eqref{eq:ql} before being pushed onto the blockchain using the SC update mechanism illustrated in Algorithm \ref{alg:scalg}.

\subsection{Data and Benchmarking Techniques}\label{sec:dbt}
For our case study, we used the NYC taxi data set \cite{NYCTaxi} for our experiments. Specifically, we randomly chose 2 million rides pertaining to May 2018 which was divided into two equal parts to denote the training and testing data sets. Restricting the rides specifically for the area of lower Manhattan resulted in approximately a little more than half million rides each in training and test data sets. The training set was used to assign rides to taxis participating in FL. 

On the basis of the test set, we determine 50 taxi trajectories which form a benchmark for FL tasks based on work done in \cite{markovTaxi}. Each trajectory comprises of 50 rides and assumes idling in case no ride is found. The sum total of fares accrued from the 50 benchmark trajectories is referred to as the Aggregated Simulation Revenue (ASR) which forms the \textit{No Learning (NL)} baseline for our case study.

The benchmark trajectories and the accompanying simulation procedure are also used to calculate ASR values for various DRL models as well. However in this case, instead of hovering in the same location upon not finding a ride, the DRL model in question is used to determine a new location to transition into \cite{markovTaxi}. The sum total of fares from the ensuing trajectories denotes the ASR value for the DRL model being considered. For robustness purposes, we perform this simulation multiple times for any DRL model and report the average ASR value.

We derive a RandomDFL mechanism that is inspired by the work done in \cite{shokri2015privacy} that can be directly applied for orchestrating a naive aggregator free FL approach. RandomDFL is described in detail in Section \ref{sec:randomDFL}

\section{Experiments}\label{sec:exp}
In order to evaluate the efficacy of BAFFLE, we focus on four key experiments. We perform a benchmark study where we compare the potential benefits from BAFFLE with respect to classical FL as well as other non FL paradigms. Next, we examine the trends arising from varying number of chunks as well as budget sizes of user devices. We then move onto a scalability analysis that demonstrates the impact of varying the total number of active user devices on the model quality. Lastly, we demonstrate the robustness of BAFFLE to the participation level (PL) parameter of BAFFLE. Further, we also show superior computational performance of BAFFLE compared to the best possible aggregator driven approach inspired by the current state-of-the-art. 


\subsection{Experimental Setup}\label{sec:expsetup}
BAFFLE was implemented and evaluated on a private Ethereum blockchain setup exclusively for our computational experiments. We employed \texttt{go-ethereum}, an official \texttt{go} based implementation of the Ethereum protocol \cite{geth} to orchestrate our private blockchain comprising of 16 Ethereum nodes. Proof-Of-Authority was used as the primary consensus protocol for all our experiments. The SC layer was developed using the Solidity programming language and deployed on the private blockchain using \texttt{go-ethereum}. The private blockchain was deployed on an Intel Xeon CPU with a clock rate of 2.40 GHz with 16 cores and 2 threads per core. We used \texttt{OpenMPI} \cite{openmpi} in conjunction with \texttt{mpi4py} \cite{mpi4py} to spawn multiple distributed memory client processes intended to simulate the user devices on the field. Each client process was assigned to a single core  of an Intel i7 CPU consisting of 12 cores. We used a 2 layer DNN with 500 perceptrons in each layer for our experiments on Keras \cite{chollet2015keras} with a TensorFlow \cite{tensorflow2015-whitepaper} backend.
\subsection{Benefits Study}\label{sec:expben}
\begin{table}[!htb]
\centering
\caption{Benefit Analysis}
\label{tab:ben}
\begin{tabular}{c|cc}
\toprule
Category   & ASR (USD)          & Benefit (\%)       \\ \hline
No Learning (NL)  & 13387.31 & -  \\ \hline
Local Learning (LL)  & 16106.02 & 20.31 \\ \hline
Classical FL (CFL)  & 18495.94 & 38.16 \\ \hline
\textbf{BAFFLE} & \textbf{18442.21} & \textbf{37.75} \\ \hline
\bottomrule
\end{tabular}
\end{table}
In this experiment we compare the benefits accrued by drivers participating in BAFFLE with respect to two other types of learning paradigms. The first comprises of a \textit{Local Learning (LL)} mechanism, wherein no model aggregation is involved. The second paradigms pertains to an aggregator driven Classical FL(CFL) scheme. We considered each taxi having accumulated approximately 700 rides in each round for a total of 50 rounds. For the FL cases we considered 16 taxis whereas for the LL case, we considered a single taxi. Table \ref{tab:ben} presents the results with respect to LL, CFL and BAFFLE mechanisms in terms of their ASR value and benefit relative to the NL baseline.

The trends depicted in Table \ref{tab:ben} provide numerous key insights into the performance of BAFFLE. Primarily, we observe that BAFFLE is able to provide a benefit of approximately 38\% which rivals the CFL approach. Further, we observe that BAFFLE and CFL approaches improve driver benefit by close to 18\% as compared to the LL case. Overall, the results demonstrate that BAFFLE is highly capable of delivering good quality machine learning models in an aggregator free, decentralized fashion.
\subsection{Sensitivity Analysis}\label{sec:expsen}
\begin{figure*}[!htb]
    \centering
    \subfigure[Gas Costs]{\includegraphics[width=0.37\textwidth,keepaspectratio]{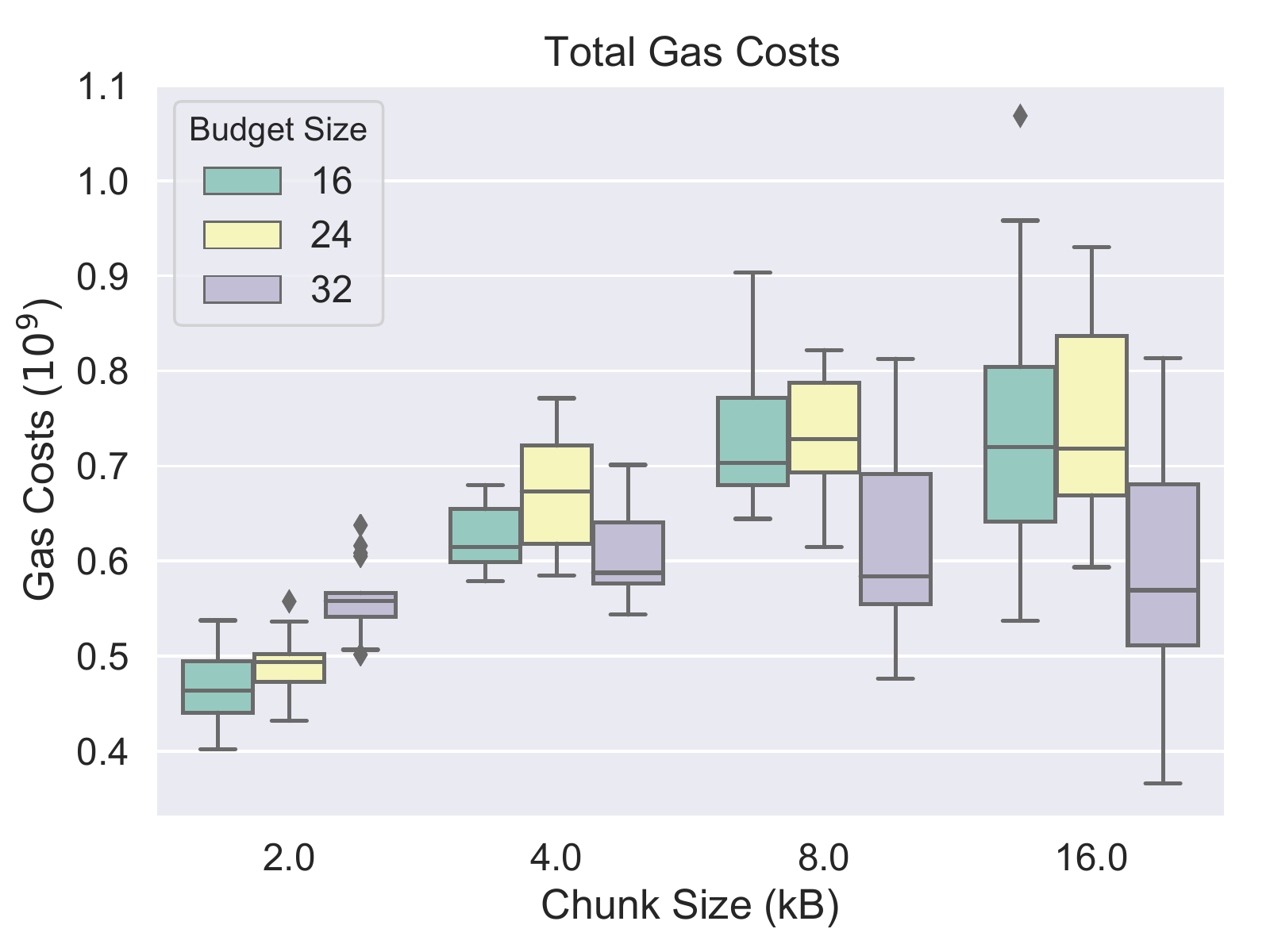}\label{fig:sengasc}}
    \subfigure[Push Time]{\includegraphics[width=0.37\textwidth,keepaspectratio]{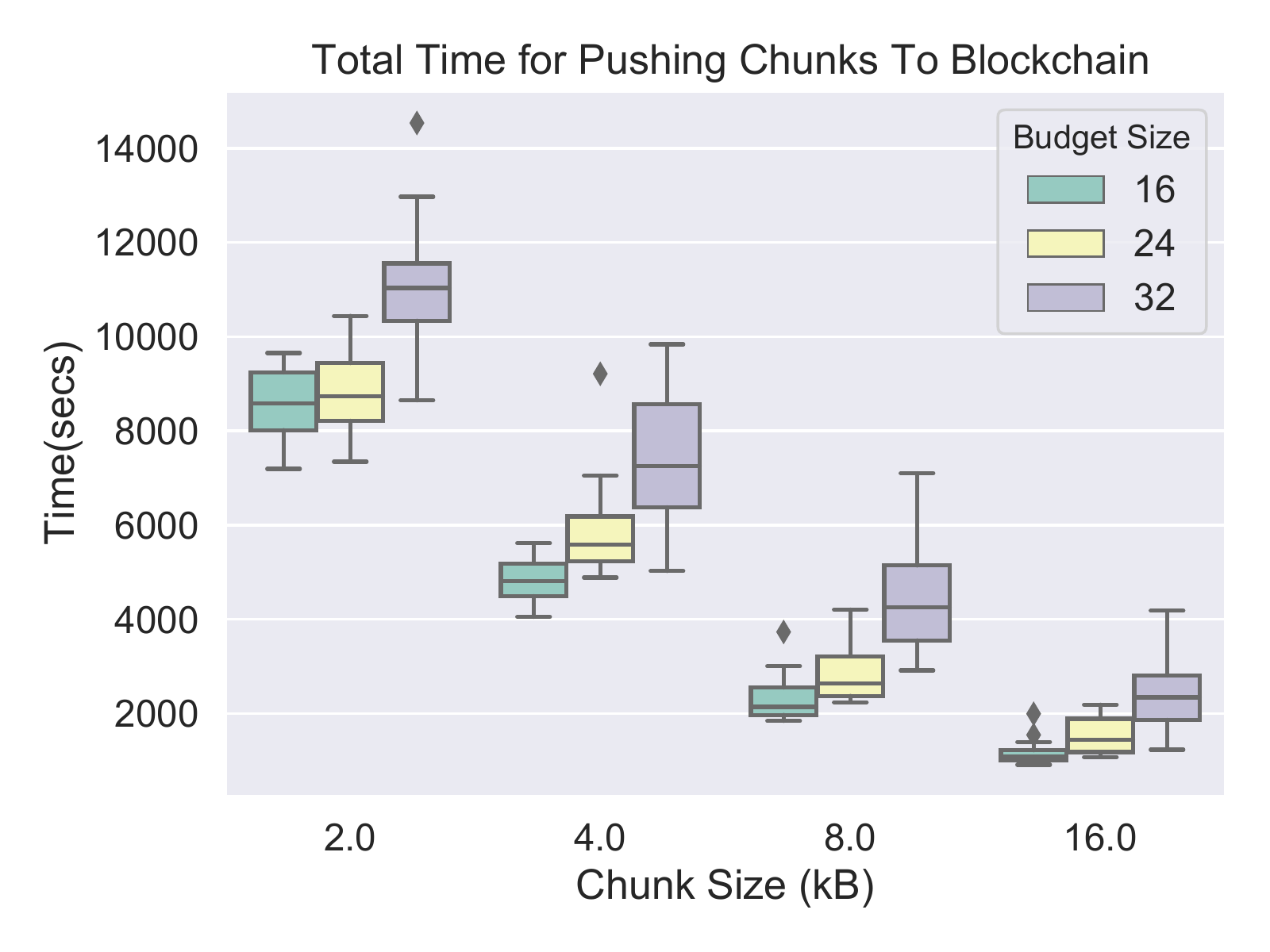}\label{fig:senpusht}}
    \caption{Performance analysis with respect to Chunk Size and Budget}
    \label{fig:chunk_sense}
\end{figure*}
\begin{table}[!htb]
\centering
\caption{Final Benefit (\%) based on Average ASR}
\label{tab:senben}
\begin{tabular}{c|c|ccc}
\toprule
Chunk  & No. Of &\multicolumn{3}{c}{Budget Size} \\\cline{3-5}
Size (kB)& Chunks  & 16          & 24          & 32          \\ \hline
2  & 738 &38.32 & 38.18 & 36.51 \\ \hline
4  & 356 &36.37 & 36.87 & 39.17 \\ \hline
8  & 181 &40.23 & 34.79 & 38.11 \\ \hline
16 & 88 &39.07 & 38.82 & 38.02 \\ \hline
\bottomrule
\end{tabular}
\end{table}

\begin{table}[!htb]
\centering
\caption{Average Total Training Time(in secs) (Std Dev.)}
\begin{tabular}{c|ccc}
\toprule
Chunk  &  \multicolumn{3}{c}{Budget Size} \\\cline{2-4}
Size (kB)   & 16          & 24          & 32          \\ \hline
2  & 87.48(2.89) & 85.97(3.26) & 73.49(2.70) \\ \hline
4  & 79.92(3.18) & 77.63(2.83) & 73.22(3.87) \\ \hline
8  & 74.16(3.70) & 71.90(2.53) & 69.79(3.09) \\ \hline
16 & 73.44(4.01) & 76.39(3.37) & 71.79(3.51) \\ \hline
\bottomrule
\end{tabular}
\label{tab:sentt}
\end{table}

We perform a robustness study to analyze the impact of variation in chunk sizes as well as local budget sizes on the overall model quality. For this experiment, we considered a total of 64 taxis, with each taxi having accumulated approximately 70 rides in each round for 125 rounds overall.
Table \ref{tab:senben} shows the benefit percentage calculated for varying chunk and budget sizes. Figure \ref{fig:chunk_sense} represents the overall trends with Figures \ref{fig:sengasc}, \ref{fig:senpusht} depicting the boxplots pertaining to Gas Costs, Push Time respectively. Table \ref{tab:sentt} shows the mean and standard deviation with respect to the training time incurred by the individual agents. 

The results for all the combinations in Table \ref{tab:senben} depict benefits that closely mirror that of the CFL approach shown in Table \ref{tab:senben} on the same training set.   Therefore, on the basis of data presented in Table \ref{tab:senben} one can conclude that BAFFLE is significantly resilient to varying degrees of budget and chunk sizes. 

On the basis of Table \ref{tab:sentt}, we conclude that time incurred for training is marginal compared to the push time depicted in Figure \ref{fig:senpusht} for all combinations of budget and chunk sizes. The relatively small training time implies that reducing the total push time is critical in ensuring a computationally efficient performance for a blockchain based FL mechanism. We can therefore draw upon the trends shown in Figure \ref{fig:chunk_sense} to reveal numerous key insights which elucidate the high computational efficiency of BAFFLE. 

Primarily, in Figure \ref{fig:sengasc} we observe a smaller variation in gas costs for the 2 kB chunk size irrespective of budget sizes. However, as the chunk size increases we see the variation in gas costs also increase substantially for all budget sizes. Second, despite the increased variation, the mean gas cost appears to saturate for higher chunk sizes. We also observe that for the budget size of 32 after the initial uptick there is a relatively more pronounced downward trend for higher chunk sizes. 
This trend can be clearly attributed to the scoring and bidding mechanism incorporated in BAFFLE. Since a higher chunk size implies lesser number of chunks, there is relatively more competition among user devices to update the same set of chunks. As a result for higher chunk sizes, only user devices which are able to consistently contribute higher scoring chunks will incur a higher gas cost. Therefore, owing to its underlying scoring and bidding mechanism, BAFFLE is able to achieve significant savings in gas costs for the users. Lastly, we observe that in Figure \ref{fig:senpusht} despite the budget size increasing, the total push time increases only marginally owing to the scoring and bidding mechanisms. Therefore, we can safely say that BAFFLE is successfully able to circumvent the computational bottleneck posed by the push step of BAFFLE .
\subsection{Scalability Analysis}\label{sec:expscal}
\begin{table}[!htbp]
\centering
\caption{Scalability analysis with varying No. of Taxis}
\label{tab:scal}
\begin{tabular}{c|cc}
\toprule
Taxis   & Average ASR (USD)  & Benefit (\%)       \\ \hline
16  & 14489.59 & 8.2  \\ \hline
32  & 16547.20 & 23.6 \\ \hline
64  & 18266.72 & 36.44 \\ \hline
128 & 18414.48 & 37.55 \\ \hline
\bottomrule
\end{tabular}
\end{table}
\begin{figure*}[!h]
    \centering
    \subfigure[Gas Costs]{\includegraphics[width=0.37\textwidth,keepaspectratio]{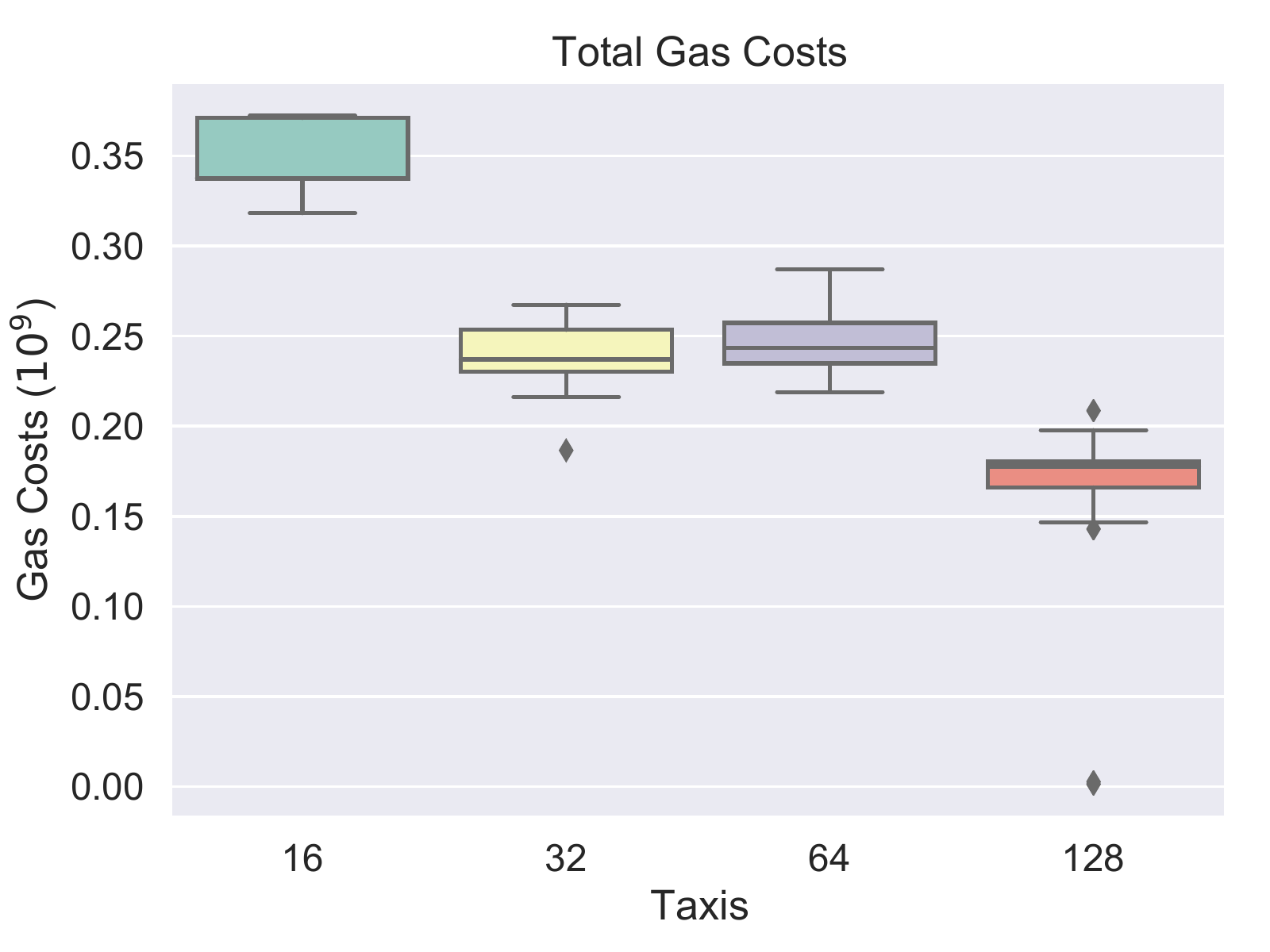}\label{fig:scalgas}}
    \subfigure[Push Time]{\includegraphics[width=0.37\textwidth,keepaspectratio]{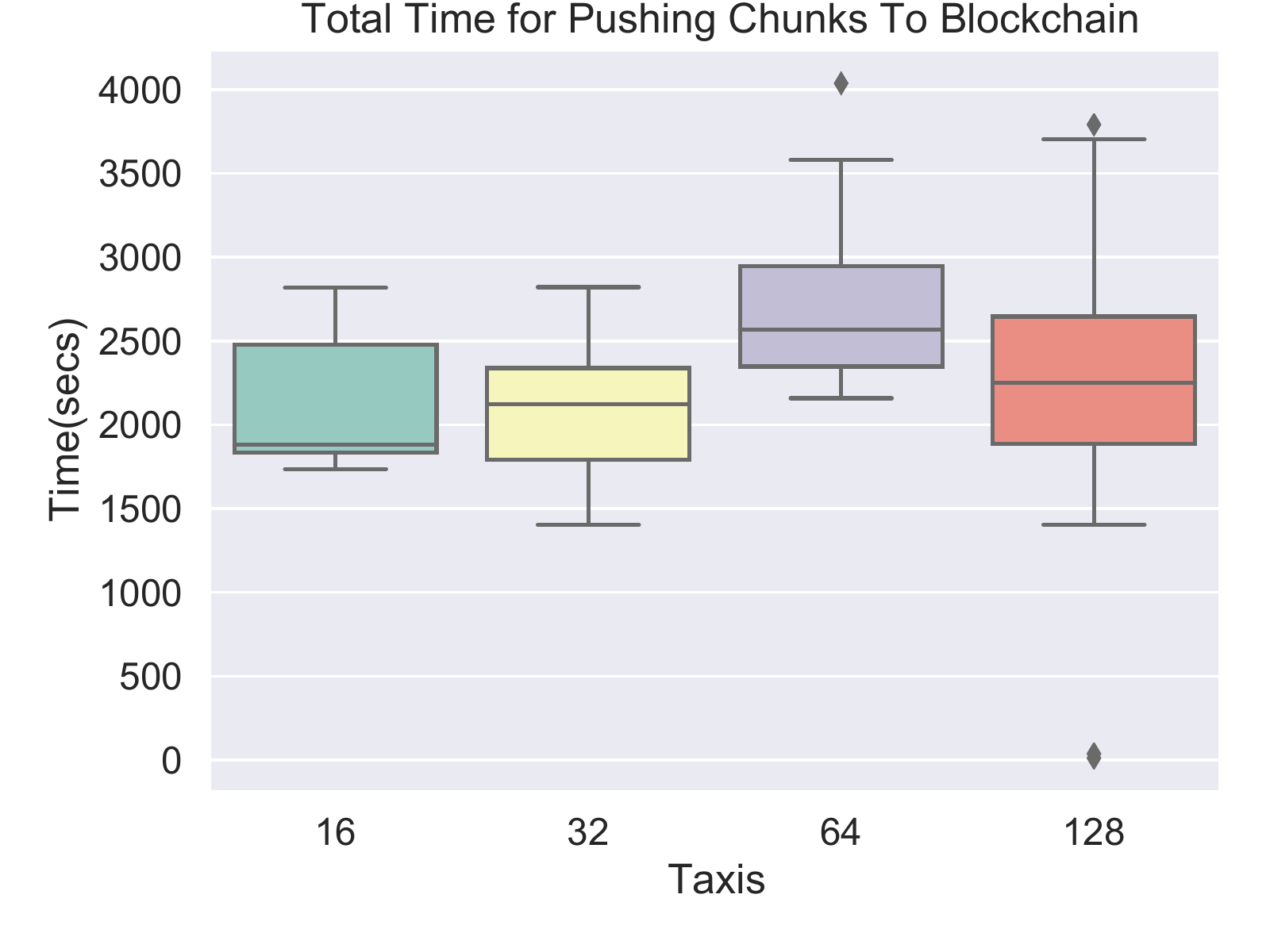}\label{fig:scalpush}}
    \caption{Weak Scaling trends}
    \label{fig:scalt}
    \vspace{-5mm}
\end{figure*}
We attempt to gauge the impact of the total number of active user devices on the performance of BAFFLE. For this experiment, we assumed each taxi having accumulated approximately 70 rides in each round for 62 rounds overall.
Table \ref{tab:scal} represents the ASR value and the ensuing benefit percentages for 16, 32, 64 and 128 taxis respectively. From the trends presented in Table \ref{tab:scal} it is apparent that increasing number of user devices results in a sizeable improvement in the model quality. However, the trends in Table \ref{tab:scal} also reveal that the improvement in model quality eventually saturates with increasing active devices potentially indicating a convergence to a globally superior model.

Figure \ref{fig:scalt} depicts box plots pertaining to the gas costs as well as the push time with varying active user devices in Figures \ref{fig:scalpush} and \ref{fig:scalgas} respectively. Figure \ref{fig:sengasc} shows a reduction in gas costs with increasing number of active devices. However, Figure \ref{fig:scalpush} reveals little variation of push time with increase in active devices. The reduction in gas costs in Figure \ref{fig:scalgas} can be attributed to greater competition arising from an increase in total number of devices. Moreover, owing to a constant push time depicted in Figure \ref{fig:scalpush} we infer that increase in number of participants effectively leads to reduction in gas costs in BAFFLE.
\subsection{Participation Level (PL) Analysis}\label{sec:exppl}
\begin{figure*}[!ht]
    \centering
    \subfigure[Model Quality Progression]{\includegraphics[width=0.32\textwidth]{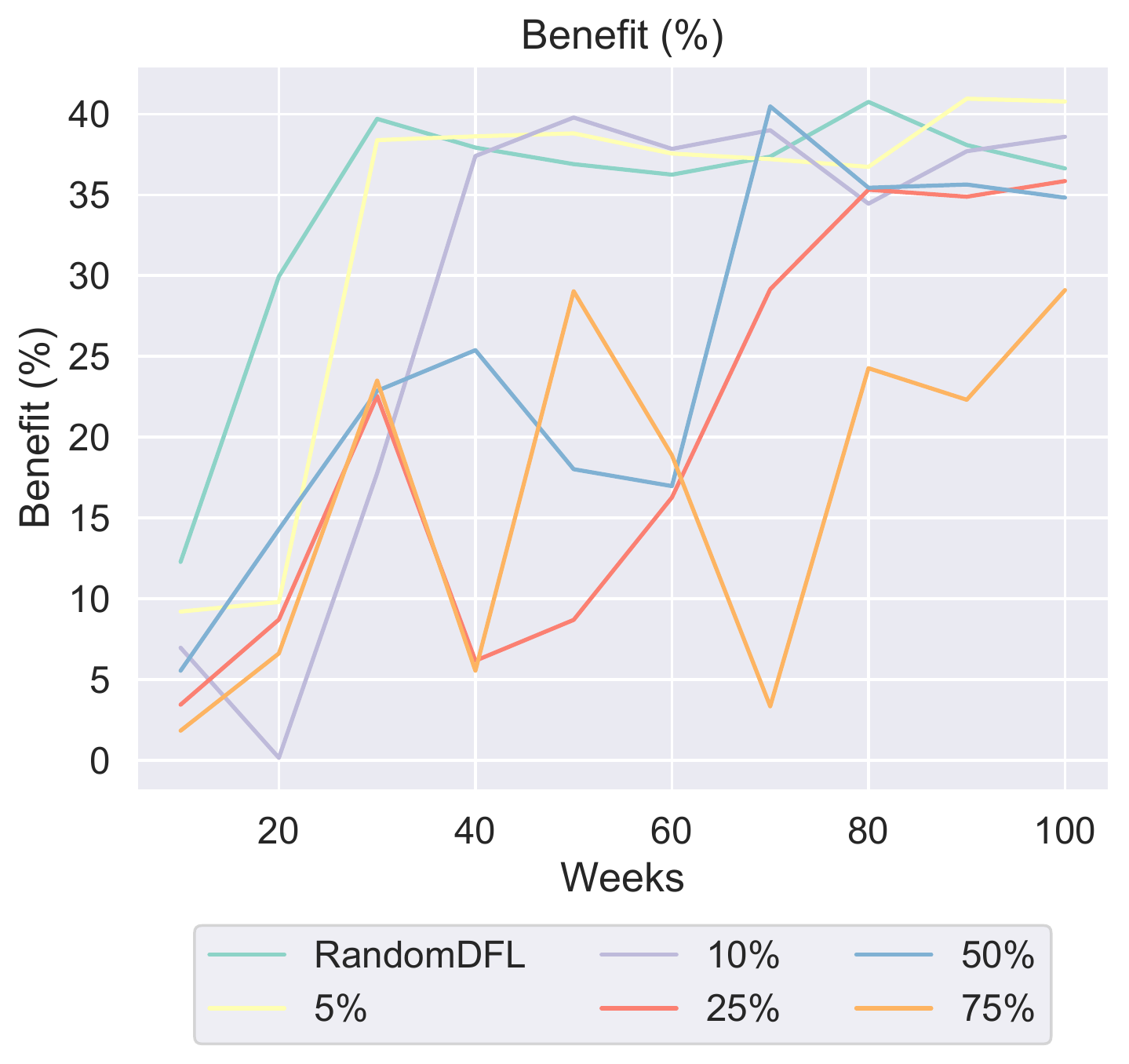}\label{fig:plmq}}
    \subfigure[Gas Cost Growth]{\includegraphics[width=2in,height=2.05in]{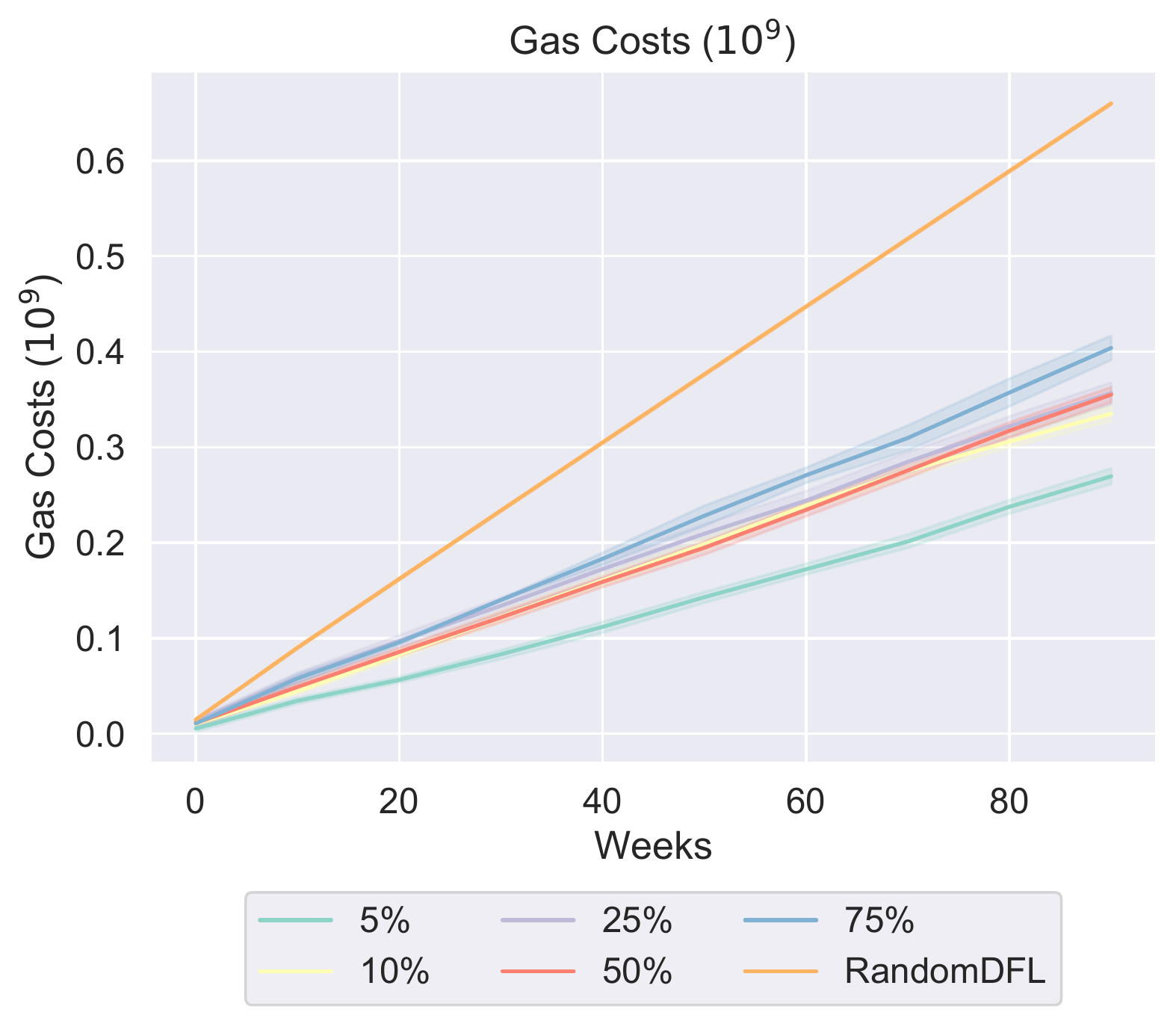}\label{fig:plgas}}
    \subfigure[Total Push Time]{\includegraphics[width=2.5in,height=2in]{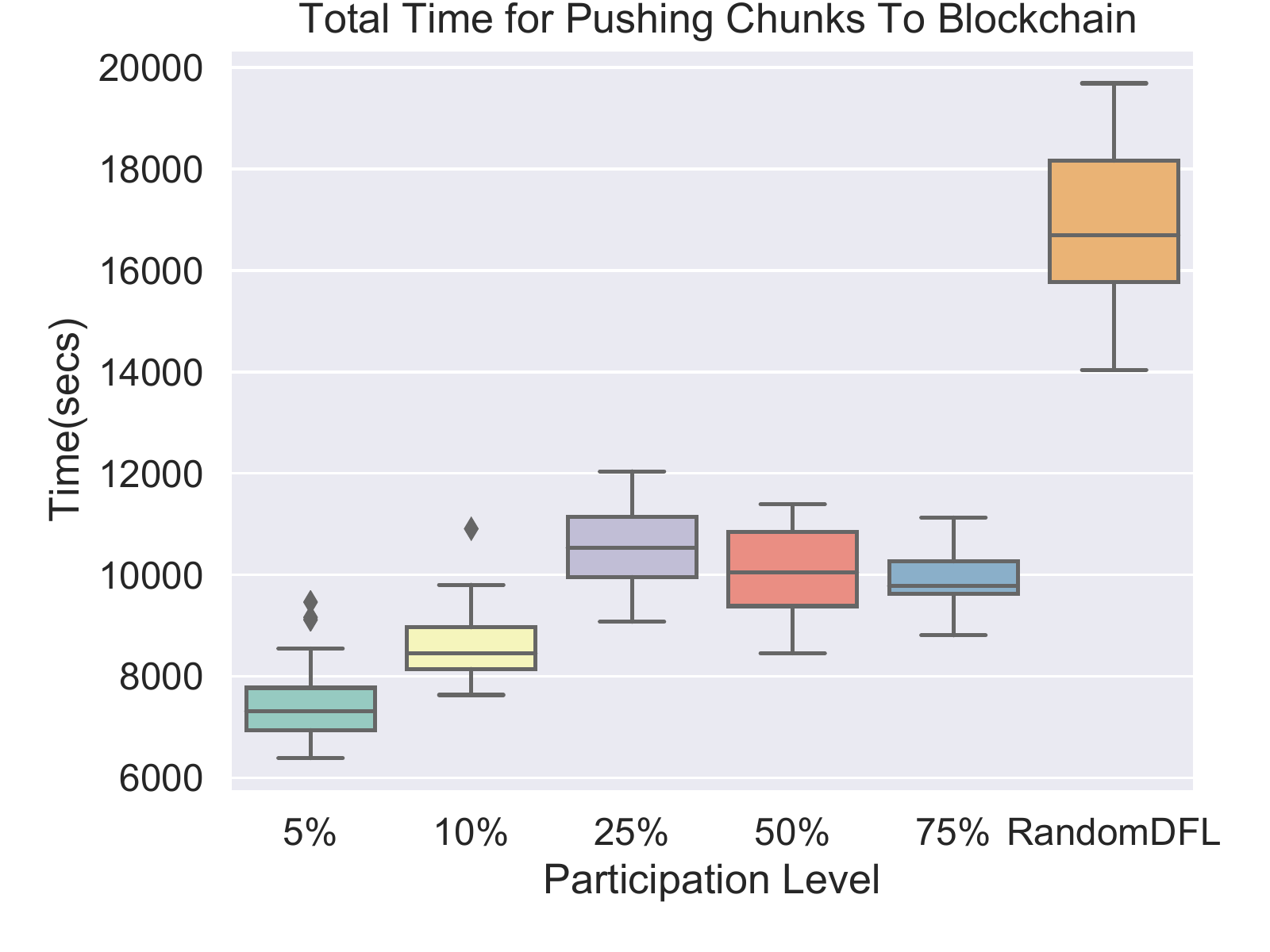}\label{fig:plpusht}}
    \caption{Performance analysis with respect to Participation Level (PL)}
    \label{fig:pl}
\end{figure*}
In this experiment we study the impact of varying the PL on the performance of BAFFLE with 64 taxis, approximately 70 rides per round and a total of 62 rounds with a budget of 16 chunks. Figure \ref{fig:pl} presents results pertaining to PL values ranging from 5\% to 75\%. Further, we also compare the RandomDFL case in which devices update the global copy without any global coordination. Figures \ref{fig:plmq}, \ref{fig:plgas} and \ref{fig:plpusht} represent the trends pertaining to the growth in model quality, gas costs and box plots of total push time pertaining to varying PL in every round. 

From Figure \ref{fig:plmq}, we observe that the fastest convergence of the model quality occurs in case of the RandomDFL case. However, the convergence characteristics of BAFFLE with a 5\% PL value closely mirrors the RandomDFL case. Overall, the trends in Figure \ref{fig:plmq} generally indicate that a lower PL value leads to a faster convergence. Figure \ref{fig:plgas} shows that a lower PL value in BAFFLE incurs a lower gas cost as well. Trends similar to Figure \ref{fig:plgas} are also exhibited in Figure \ref{fig:plpusht} wherein a lower PL value in BAFFLE corresponds to a lower total push time as well. We observe that in general, BAFFLE incurs barely half the gas cost and push time as compared to the RandomDFL case. In fact, BAFFLE outperforms the RandomDFL case by a factor of more than 2 with a PL value of 5\% both in terms of the gas cost as well as the push time. Since fewer devices are pushing to the global model copy every round, the chances of multiple devices pulling the same global model is significantly higher in case of a lower PL value. This leads to greater stability in the decentralized process which ultimately leads to a faster convergence for a low PL value as shown in  Figure \ref{fig:plmq}. 

BAFFLE incurs significantly lower gas costs compared to the RandomDFL case owing to minimization of redundant updates. Due to the decentralized round delineation and a robust scoring and bidding process, devices only push chunks that are among the best in the round. As a result, collision among devices for the same chunk is completely eliminated leading to a much lower gas cost and push time.

\section{Conclusion and Future Work}\label{sec:conc}
In this paper we investigate the use of the blockchain for realizing a decentralized aggregator free FL mechanism. We design and develop BAFFLE, a custom made blockchain based framework for aggregator free FL. In our framework, we successfully eliminate the role of a centralized aggregator by effectively decentralizing the concepts of round delineation, user device selection and model aggregation with the help of an SC. Further, in order to circumvent the computational restrictions imposed by the blockchain, we employ an effective model partitioning and serialization mechanism that enables independent and parallel model updates. We orchestrate BAFFLE on a private Ethereum blockchain network with a Solidity driven SC implementation. We argue that the operational and computational benefits of aggregator free FL has significant potential for solving  business problems for micro scale enterprises. We support our claims by applying BAFFLE to a case study pertaining to the ride sharing and taxi industry which serves as a perfect example of a micro scale enterprise. Our case study utilizes the BAFFLE framework to improve driver revenue based on a DRL model that is collectively augmented by all drivers using FL. We show that BAFFLE yields approximately a 40\% improvement in driver revenues compared to non FL approaches. We further show that despite being aggregator free, BAFFLE's result quality matches that of classical FL schemes that require investment in an aggregator. Moreover, BAFFLE performs significantly better compared to other aggregator free approaches that are inspired by the current state of the art. 

Our work shows that an aggregator free approach to FL offers significant potential for revolutionizing small scale organizations and their businesses by delivering quality machine learning models at lower costs. Driven by a robust decentralized platform like the blockchain, the benefits of FL could impact a variety of domains leading to widespread adoption. The issue of aggregator free FL opens up new avenues for research especially in the blockchain domain. Therefore, our future work is driven by the desire to incorporate other machine learning paradigms as well as differential privacy into BAFFLE. Incorporating an aggregator free FL for complex paradigms like CNNs and LSTMs will go a long way to enable wider adoption of FL. 


\small
\bibliography{main}
\bibliographystyle{ieeetr}

\appendix
\section{Appendix}
\subsection{Centralized Deep Batch Q Learning}\label{sec:cdbq}
Algorithm \ref{alg:c} details the centralized batch Deep Q Learning for improving driver revenue. It starts with observation of a new ride set every epoch. For every ride in the ride set, the existing Q-value estimate is updated with the fare collected for the ride and a discounted future reward. The discounted future reward is based on the action that gives highest Q value originating from the destination state. Based on the observed set of rides, a Deep Neural Network (DNN) is used to calculate the next Q function estimate.
\begin{algorithm}
\caption{Centralized Deep Neural Fitted Q}\label{alg:c}
\begin{algorithmic}
\For{$k=0 \ldots$}
\State observe new ride set $\mathcal{H}^k$
\State pull latest available model $Q$ from blockchain
\State perform averaging $Q_k \leftarrow \frac{Q^j_k+Q}{2}$
\State update $\tilde{Q}^j_k$ based on Equation \eqref{eq:qbellman}
\State locally train $Q^j_{k+1}$ via Equation \eqref{eq:qsgd}
\EndFor
\end{algorithmic}
\end{algorithm}
\subsection{Random Decentralized FL (RandomDFL)}\label{sec:randomDFL}
\begin{algorithm}
\caption{Randomized Decentralized Deep Neural Fitted Q}\label{alg:r}
\begin{algorithmic}
\For{taxi: $j=1\ldots P$}
\For{$k=0 \ldots$}
\State observe new ride set $\mathcal{H}^k$
\State pull latest available model $Q$ from blockchain
\State perform averaging $Q_k \leftarrow \frac{Q^j_k+Q}{2}$
\State update $\tilde{Q}^j_k$ based on Equation \eqref{eq:qbellman}
\State locally train $Q^j_{k+1}$ via Equation \eqref{eq:qsgd}
\State push random set of chunks $C^k \subseteq \mathcal{C}, |C^k| = B$
\EndFor
\EndFor
\end{algorithmic}
\end{algorithm}
In the randomized version represented in Algorithm \ref{alg:r}, the SC is considered to be naive. User devices are free to update any chunks subject to their own budget values. In this naive randomized version, some chunk updates are bound to get wasted owing to the fact that they may be overwritten by another user device's contribution before the previous update has had a chance to be read by the other agents.

\end{document}